%% file: lower_bound_main.tex
\documentclass{llncs}
\usepackage{amsmath}
\usepackage{amssymb}

\usepackage{tikz}

\usepackage[linesnumbered,vlined]{algorithm2e}
\usepackage{color}

\usepackage{verbatim}

\definecolor{defseagreen}{cmyk}{0.69,0,0.50,0}

\newcommand{\smallerLEQ}{\leq} 
\newcommand{\smallerNEQ}{<} 
\newcommand{\union}{\cup}
\newcommand{\intersection}{\cap}
\let\implies=\undefined%
\newcommand{\implies}{\Rightarrow}

\newcommand{\mmodels}{\models_\mathrm{min}}
\newcommand{\comprehension}[2]{\ensuremath{\left\{ {#1} \;|\; {#2}\right\}}}
\newcommand{\powerset}{{\mathcal P}}
\newcommand{\circumscription}{\text{CIRC}}

\newcommand{\GCWA}{\text{GCWA}\xspace}
\newcommand{\EGCWA}{\text{EGCWA}\xspace}
\newcommand{\ECWA}{\text{ECWA}\xspace}

\newcommand{\cwrule}[3]{\noindent{\bf #1}~{\bf (#2)}: #3\\[-0.1cm]}

\newcommand{\mathematics}[1]{$#1$}

\def\entailsMinimal {{\sc Entails-Min}\xspace}
\def\negationFree{{\sc Free-For-Negation}\xspace}
\def\negationFreeAll {{\sc Free-For-Negation-All}\xspace}

\def\theTitle{Counterexample Guided Abstraction Refinement Algorithm for Propositional Circumscription}
\def\theTitleBroken{\theTitle}

\def\miko{Mikol\'a\v{s} Janota}
\def\jpms{Joao Marques-Silva}
\def\radu{Radu Grigore}
\title{\theTitleBroken}
\subtitle{\it This article is an extended version of an article of the same name accepted to JELIA 2010}
\author{{\miko}\inst{1}\and{\radu}\inst{2}\and {\jpms}\inst{3}}
\institute{%
INESC-ID, Lisbon, Portugal\and
Queen Mary, University of London\and
University College Dublin, Ireland%
}

\definecolor{citeblue}{rgb}{0.1,0,.25}
\definecolor{refcolor}{rgb}{0,0,0.4}
\usepackage[pdftex,a4paper=true,colorlinks=true,bookmarks=true,%
           linkcolor=citeblue,citecolor=citeblue,
            urlcolor=blue,%
            plainpages=false]{hyperref}
\hypersetup{ 
pdftitle={{\theTitle}},
pdfauthor={{\miko}}
}

\def\algorithmautorefname {Algorithm}
\begin {document}
\maketitle
\input {lower_bound}
\input {conclusion}

\bibliographystyle{splncs03}
\bibliography{refs,proceedings}
\clearpage
\appendix
\input {proofs_mj}
\input {3_sets}
\end {document}

%% file: lower_bound.tex

\begin {abstract}
Circumscription is a representative example of a non\-mono\-to\-nic
reasoning inference technique. Circumscription has often been studied
for first order theories, but its propositional version has also been
the subject of extensive research, having been shown equivalent to
extended closed world assumption (ECWA). Moreover, entailment in
propositional circumscription is a well-known example of a decision
problem in the second level of the polynomial hierarchy. This paper
proposes a new Boolean Satisfiability (SAT)-based algorithm for
entailment in propositional circumscription that explores the
relationship of propositional circumscription to minimal models. The
new algorithm  is inspired by ideas commonly used in SAT-based model
checking, namely counterexample guided abstraction refinement. In
addition, the new algorithm is refined to compute the theory closure
for generalized close world assumption (GCWA). Experimental results
show that the new algorithm can solve problem instances that other 
solutions are unable to solve.
%
%
\end {abstract}

\input{intro}

\input{prelim}
\input{probs}

\section {Computing \entailsMinimal} \label{sec:emin}

The algorithm we wish to develop will be using a SAT solver.
This gives us two objectives.
One objective is to construct a propositional formula that corresponds to the validity of $\phi\mmodels\psi$.
The second objective is to avoid constructing an exponentially large formula.
%
We begin by observing that if $\phi\mmodels\psi$ is to hold, then any model of $\phi$ that violates $\psi$ must {\em not} be a minimal model.

\begin {proposition}\label {proposition:main}
$\psi$ holds in all minimal models of $\phi$ iff
 any model $\nu$ of $\phi$ where $\lnot\psi$ holds is not a minimal model of $\phi$.
\[\left [\phi\mmodels\psi\right] 
 \Leftrightarrow 
\left[ 
(\forall \nu) \left((
\nu\models\phi\land\lnot\psi)
\Rightarrow (\exists \nu')(\nu' \smallerNEQ \nu \land \nu'\models \phi) \right)
\right]
\]
\end {proposition}

\autoref {proposition:main} tells us  that whether $\phi\models_{\textrm{min}}\psi $ holds or not can be decided
by deciding whether the following formula is valid:
\begin {equation}\label{equation:first_order}
 (\forall \nu)\left((\nu\models \phi\land\lnot\psi)
               \Rightarrow (\exists \nu')(\nu' \smallerNEQ \nu \land \nu'\models \phi)\right)
\end {equation}

Since our first objective is to find a propositional formula, we need to eliminate $\cdot\models\cdot$ and quantifiers from~\eqref{equation:first_order}.
First, let us focus on the subformula $(\exists \nu')(\nu' \smallerNEQ \nu \land \nu'\models \phi)$, which expresses that $\nu$ is not a minimal model.

\begin {proposition}\label {proposition:substitution}
A model~$\nu$ of~$\phi$ is \emph{not} minimal iff there exists a set~$S$ of variables such that $\nu$~is a model of $\phi[S\mapsto0]$, and $\nu(x)=1$ for some~$x\in S$.
%
\begin {equation}\label {equation:substitution}
(\exists \nu')(\nu' \smallerNEQ \nu \land \nu'\models \phi) 
\Leftrightarrow
(\exists S\subseteq V)\left(\nu\models \phi[S\mapsto 0]\land(\exists x\in S)(\nu(x)=1)\right)
\end {equation}
\end {proposition}

\begin {example}
Let $\phi =\lnot x\lor y$. 
The model $\mu=\{x^0,y^0\}$ is minimal 
and the right-hand side of \eqref{equation:substitution} is invalid since 
there is no set $S$ satisfying the condition $(\exists x\in S)(\nu(x)=1)$.
Let   $\nu =\{x^0,y^1\}$ and
let us choose $S=\{x,y\}$, which yields
$\phi[S\mapsto 0] = 1$. 
$\nu$ is {\em not} minimal and the right-hand side of \eqref{equation:substitution} is valid since $\nu\models 1$ and $\nu(y)=1$.
\end {example}

Replacing the left-hand side of \eqref{equation:substitution} with the right-hand side of \eqref{equation:substitution} in \eqref{equation:first_order} yields the following formula:
\begin {equation}\label{equation:first_order1}
 (\forall \nu)\left((\nu\models \phi\land\lnot\psi)
               \Rightarrow 
(\exists S\subseteq V)\left(\nu\models \phi[S\mapsto 0]\land(\exists x\in S)(\nu(x)=1)\right)
\right)
\end {equation}

Removing the universal quantifier  and replacing          
existential quantifiers with the Boolean operator                
$\lor$ in \eqref{equation:first_order1}, 
gives us that \eqref{equation:first_order1} holds iff 
the following formula is a tautology:
\begin {equation}\label {equation:propositional}
(\phi\land\lnot\psi)\Rightarrow
\bigvee_{S\in \powerset(V)} \left (\phi[S\mapsto 0]\land  \bigvee_{x\in S}x  \right)
\end {equation}

Intuitively, \eqref{equation:propositional} expresses that if $\psi$ is violated in a model of $\phi$,
then a different model of~$\phi$ is obtained by flipping a set of variables to~$0$.
That this model is indeed different is guaranteed by the  condition $\bigvee_{x\in S}x$. 
The model obtained by the flipping serves as a \emph{witness} of 
that the model violating $\psi$ is \emph{not} minimal.

If \eqref{equation:propositional} is constructed, its validity can be decided by calling a SAT solver on its negation.
However, the formula is too large to construct since it requires considering all subsets of $V$. 
Therefore, we construct a stronger version of it that considers only {\em some} subsets of $V$.
This stronger version is referred to as the \emph{abstraction} of \eqref{equation:propositional} 
and always has the following form:
\begin {equation}\label {equation:abstraction}
(\phi\land\lnot\psi)\Rightarrow
\bigvee_{S\in W} \left (\phi[S\mapsto 0]\land  \bigvee_{x\in S}x  \right)
\qquad
\text{ where  $W\subseteq\powerset(V)$}
\end {equation}

Each abstraction is determined by a set of sets of variables $W$.
For any $W$, if the abstraction \eqref {equation:abstraction} is shown to be a tautology, then \eqref{equation:propositional} is also a tautology
and we are done because we have shown that $\phi\models_\textrm{min} \psi$. 
If the abstraction is not a tautology, it is either because $\phi\models_\textrm{min} \psi$ does not hold
or the abstraction is overly strong---it is too coarse. If the abstraction is shown to be too coarse, a different abstraction must be considered.

\begin {example}
Let us show that \hbox{$\lnot x\lor y\mmodels\lnot y$}.
First, let us try $W_1 =\{\{y\}\}$, which yields the abstraction
$((\lnot x\lor y)\land y) \implies \lnot x$.
This abstraction is not a tautology. In particular, it is violated by the assignment $\{x^1,y^1\}$, 
which means that flipping $y$ to value 0 in this assignment does not yield a model.
Now, let us try $W_2 =\{\{x, y\}\}$, which yields the abstraction
$((\lnot x\lor y)\land y) \implies 1$. 
This abstraction is a tautology, which means that any model where $y$ is 1 can be turned into another model by flipping both $x$ and $y$ to 0.
Therefore,  $\lnot x\lor y\mmodels \lnot y$.
\end {example}

\begin {example}
Let \hbox{$\phi =\lnot x\lor \lnot y\lor\lnot z$} 
and  \hbox {$\psi=
(\lnot x\lor \lnot y)\land
(\lnot x\lor \lnot z)\land
(\lnot z\lor \lnot y)$}
  Let us show that $\phi\mmodels\psi$.
  Let  us choose  the abstraction defined by the set $W=\{\{x\},\{y\}\}$.
  The following diagram demonstrates that each model  violating $\psi$ has a witness corresponding to one of the sets in $W$.

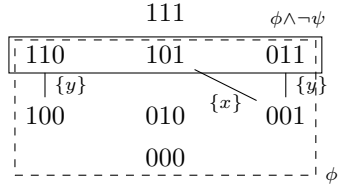
\input {example1}
\hspace{.2cm}%
  \begin{minipage}[b]{0.53\linewidth}
    \begin{small}\it
     Each triple represents a variable assignment where the elements represent the values of    $x$, $y$, and~$z$, respectively.
     Models  and their pertaining witnesses are connected by an edge, 
    which is labeled by the set of variables $S$ whose values are flipped to $0$ to obtain the witness. 
    \end {small}
  \end{minipage}
\end {example}

The  approach of searching for the right abstraction follows the
Counter-Example Guided Abstract Refinement (CEGAR)
loop~\cite{ClarkeEtAl-CAV00}.
If the abstraction is a tautology, the search terminates.
If the abstraction is not a tautology, it is weakened by adding some set of variables $S$
to the set $W$. 
This weakening is referred to as \emph {refinement} and is done by investigating the counterexample that shows that the current abstraction is not a tautology.
If it cannot be refined, \eqref{equation:propositional} is not a tautology and $\phi\mmodels\psi$ does not hold.

\begin{algorithm}[t]
\IncMargin{1em} %
\DontPrintSemicolon%
\SetKwData{out}{outc}%
\SetKwData{false}{false}\SetKwData{true}{true}
\SetKwFunction{SAT}{SAT}\SetKwFunction{smaller}{Smaller}
\SetKwInOut{Input}{input}\SetKwInOut{Output}{output}
\Input{formulas $\phi $ and $\psi $}
\Output{\true iff $\phi\models_{\textrm{min}}\psi$}
\BlankLine
\label {step:initialization} $\omega\gets \phi\land\lnot \psi$\;
\While{\true} {
\label {step:SAT_call} $(\out_1,\nu) \gets \SAT(\omega)$\;
\If {$\out_1 =\false$} {\label {step:tautology}\Return \true\tcp*[r]{no counterexample was found}} 
\label {step:get_smaller}
$(\out_2,\nu')\gets \SAT\left(\phi\land\bigwedge_{\nu(x)=0}\lnot x \land \bigvee_{\nu(x)=1}\lnot x\right)$%
\tcp*[f]{find $\nu' \smallerNEQ \nu $}\;
\If(\tcp*[f]{$\nu$ is minimal}) {$\out_2 =\false $} 
{\label {step:false} \Return \false\tcp*[f]{abstraction cannot be refined } }
 $S\gets \{  x\in V\,|\, \nu(x)=1 \land \nu'(x)=0\}$\;
\label {step:refine} $\omega\gets \omega \land  (\lnot\phi [S\mapsto 0]\lor \bigwedge_{x\in S}\lnot x) $%
\tcp*[r] {refine}
}
\caption{Refining}\label{algo_refinement}
\end{algorithm}\DecMargin{1em}

\autoref {algo_refinement} realizes the idea outlined above.
The algorithm 
maintains the negation of the abstraction in variable $\omega$
and 
starts with $W$ being the empty set.
Therefore the initial abstraction is $\left (\phi\land\lnot \psi\right)\implies 0$
with the negation being $\phi\land\lnot \psi$ (\autoref {step:initialization}).
The test whether the abstraction is a tautology or not is done by calling a SAT solver on its negation (\autoref {step:SAT_call}). 
If the negation is unsatisfiable---the abstraction is a tautology---then the algorithm terminates and returns {\sf true} (\autoref{step:tautology}).
If a model $\nu$ is found showing that the abstraction is not a tautology,
it means that for any assignment that is obtained from $\nu$ by flipping some set of variables in $S\in W$ to 0 is not a model of $\phi$.
The algorithm looks for a model $\nu'$ that is strictly smaller than $\nu$ 
applying \autoref {proposition:smaller}
(\autoref {step:get_smaller}).
If there is no model strictly smaller than $\nu$ then the algorithm terminates and returns {\sf false} since $\nu$ is a minimal model and violates $\psi$ (\autoref {step:false}).
If there is a model $\nu'$ that is strictly smaller than $\nu$, there is some set of variables that are 1 in $\nu$ but are 0 in $\nu'$. This set of variables is added to the sets determining the abstraction (\autoref {step:refine}).
 Observe that a set $S$ will be used at most once to refine the abstraction
since once the set is added to $W$,  an assignment for which flipping  $1$ to  $0$  for variables in $S$ yields a model  cannot satisfy the negation of the abstraction. 
Consequently, the algorithm is terminating and will perform at most as many iterations as there are subsets of the set $V$.
\section {Computing \negationFree} \label{sec:ffn}

This section specializes \autoref{algo_refinement} to compute   variables  free for negation---variables that take value $0$  in all minimal models.
As mentioned earlier, this problem is a special case of the problem \entailsMinimal, studied in the previous section: $x$ is free for negation in $\phi $ iff $\phi\mmodels \lnot x$.
However, focusing on this type of formulas 
enables a more efficient implementation of the algorithm.

The abstractions used in the previous section
have to contain the condition that at least one of the variables being flipped to 0 is 1 to guarantee the corresponding witnesses is strictly smaller (see \eqref{equation:propositional}).
For  variables free for negation these conditions will not be needed thanks to the following proposition.

\begin{proposition}\label {proposition:dispensable_flipped}
Let $\nu$ be a model of $\phi$ s.t.\ $\nu(x) = 1$ for a variable $x$.
If $x$ is free for negation, then there exists a model $\nu'$ of $\phi$ 
s.t.\ $\nu'\smallerNEQ \nu $ and $\nu'(x)=0$.
\end {proposition}

\autoref {proposition:dispensable_flipped} tells us that if  \hbox{$\nu(x)=1$} and $x$ is free for negation,
there must be a witness $\nu'$ that flips $x$ to $0$ (and possibly some other variables).
This ensures that $\nu$ and $\nu'$ are different.
This observation enables us to compute $\phi\mmodels \lnot x$ by determining  the validity of a stronger and more concise formula than before.
\begin {proposition}\label {proposition:dispensable}
A variable $x$ is free for negation in $\phi$ iff the following formula is a tautology.
\begin {equation}\label {equation:dispensable}
(\phi \land x) \implies
\bigvee\hspace{-.45em}\raisebox{-.5em}{$\scriptstyle{S\subseteq V\land x\in S}$}\;\;\phi [S\mapsto 0]
\end {equation}
\end {proposition}

The abstraction of \eqref{equation:dispensable} is analogous to the one used in the previous section
with the difference that only sets of variables containing~$x$ are considered.
Hence, the abstraction always has the following form.
\begin {equation}
(\phi\land x)\Rightarrow
\bigvee\hspace{-.45em}\raisebox{-.5em}{$\scriptstyle S\in W$}\;\;
\phi [S\mapsto 0], 
\text { where } W\subseteq \powerset(V) \text{ and }
(\forall S\in W)(x\in S) 
\end {equation}

\subsection {Constructing and Refining Abstraction}
\begin{algorithm}[t]
\IncMargin{1em} %
\DontPrintSemicolon%
\SetKwData{out}{outc}%
\SetKwData{false}{false}\SetKwData{true}{true}
\SetKwFunction{SAT}{SAT}\SetKwFunction{smaller}{Smaller}
\SetKwInOut{Input}{input}\SetKwInOut{Output}{output}
\Input{CNF formula $\phi $ and a variable $x$}
\Output{\true iff $\phi\mmodels \lnot x$}
\BlankLine
\label {line:substitution_0}$\phi_0\gets\phi [x\mapsto 0]$\;
$\phi_0'\gets
\comprehension{\lnot r_c\lor c}{c\in \phi_0}\union
\comprehension{\lnot l \lor r_c}{c\in \phi_0, l\in c}\union
\left\{\bigvee_{c\in \phi_0} \lnot r_c\right\}$\label {line:transformation}\;
$\omega\gets \phi\land x \land \phi_0'$ \;
\While{\true} {
$(\out_1,\nu) \gets \SAT(\omega)$\;
\If {$\out_1 =\false$} {\Return \true\tcp*[r]{no counterexample was found}} 
\label {step:get_smaller}

$(\out_2,\nu')\gets \SAT\left(\phi\land\lnot x \land \bigwedge_{\nu(z)=0 }\lnot{z}\right)$%
\tcp*[f]{find $\nu' \smallerNEQ \nu$ and $\nu' (x) = 0 $}\label {line:dispensable_smaller}\;
\If {$\out_2 =\false $} 
{\label {step:false} \Return \false\tcp*[f]{abstraction cannot be refined } }
 $S\gets \{  z\in V\,|\, \nu(z)=1 \land \nu'(z)=0\}$\;
%
 $C_p\gets\comprehension {c \in \phi_0}{ \left (c\intersection S\right)\neq\emptyset }$ \tcp*[f] {clauses with some $y\in S$}\;
 $C_n\gets\comprehension {c \in \phi_0}{ \left (c \intersection \lnot S\right)\neq\emptyset }$ \tcp*[f] {clauses with some $\lnot y\in S$}\;
 $C\gets\comprehension {c'}{ c\in (C_p\smallsetminus C_n) \land c'=c[S\mapsto 0]}$ \tcp*[f] {new clauses} \;
%
 $\omega\gets \omega\union%
\comprehension{\lnot r_c\lor c}{c\in C}\union%
\comprehension{\lnot l \lor r_c}{c\in C, l\in c}$
\tcp*[f] {representation}\;
 $\omega\gets \omega\union%
\left\{%
\bigvee_{c\in\phi\smallsetminus (C_n\union C_p)} \lnot r_c \lor \bigvee_{c\in C} \lnot r_{c}%
\right\}$\tcp*[f] { negation of clauses}
}
\caption{Deciding whether a variable is free for negation}\label{algo:dispensable}
\end{algorithm}\DecMargin{1em}

Whenever the abstraction is being refined (weakened) the size of the formula representing the negation of the abstraction increases.
Since the abstraction is refined in the worst case exponentially many times,
it is warranted to pay attention to the size of the formula representing the negation of the abstraction.

The negation of an abstraction is a conjunct of the left-hand side of the implication and formulas  capturing the substitutions.
\begin {equation}\label {equation:negation}
(\phi\land x) \land %
\bigwedge
\hspace{-.05em}\raisebox{-.5em}{$\scriptstyle S\in W$}\; {\lnot\phi [S\mapsto 0]}, %
  \text { where } W\subseteq \powerset(V) \text{ and } (\forall S\in W)(x\in S) 
\end {equation}

When the abstraction is being refined, a new set of variables $S$ is added to the set $W$, therefore, 
the negation of the abstraction is strengthened by conjoining it with $\lnot\phi [S\mapsto 0]$. 
We aim to implement this strengthening without duplicating those parts of the formula that are already present.

\autoref{algo:dispensable} outlines this procedure.
Since all the sets $S$ must contain $x$, the algorithm starts with the abstraction determined by $W =\{\{x\}\} $. 
In the initialization phase,  the negation of this abstraction is $\phi\land x\land\lnot\phi[x\mapsto 0]$
and is computed using the Tseitin transformation~\cite{Tseitin68}.
Each clause $c$ in $\phi[x\mapsto 0]$ is represented by a fresh variable $r_c$
and a clause is added that expresses that at least one of these variables must be 0 (\autoref {line:transformation}).
As in the previous section,
variable $\omega$ represents the negation of the abstraction (see \eqref{equation:negation}).

When the abstraction is being refined, the formula  in variable $\omega $ is conjoined with $\lnot\phi[S\mapsto 0]$.
Since $\omega$ already contains clauses from $\lnot\phi[x\mapsto 0] $, we need to consider only those clauses that contain literals on the variables in $S$.
Clauses containing negative literals on variables from $S$ are skipped, 
positive literals are removed.
Each of the affected clauses is represented by a fresh Tseitin variable.
Finally, a clause is added to express that one of the clauses in $\phi[S\mapsto 0]$ is $0$. Note that
this clause is referring to the original Tseitin variables for the clauses that are not affected by the substitution besides the freshly created ones.
Note that when looking for a model $\nu'\smallerNEQ\nu$, the algorithm
requires that $x$ has value 0 in $\nu'$ 
since
the set $S$ must contain $x$ (\autoref {line:dispensable_smaller}).

\subsection {Finding Models}
An abstraction is refined according to two responses from the underlying SAT solver ($\nu$ and $\nu'$).
This enables us to devise heuristics that prefer some responses of the solver to another.
The motivation for these heuristics is to find abstractions where the set $W$ determining the abstraction contains few sets $S$.
Dually, this means that each of  $S\in W$ yields a witness for many models.
The  heuristics  used in the current implementation are motivated by the two following examples.

\begin {example}
Let \hbox {$\phi = (x\implies y)\land (w\lor z)$}. The abstraction defined by  $W =\{\{x,y\}\}$ shows that $\phi\mmodels\lnot y$ 
 since flipping both~$x$ and~$y$ in any model yields a model (a witness).
The abstraction determined by $W =\{\{x,y,z\}\}$ is not sufficient. 
This abstraction provides a witness for models with $w$ having value 1 but  not for the others. Intuitively, variable  $z$ is irrelevant to the relation $x\implies y$ and therefore it is better to choose a small $S$.
\end {example} 

\begin {example}
Let $\phi = x\implies (y\lor w_1\lor\dots w_n)$ and let us prove that $\phi\mmodels\lnot y$. 
The abstraction  determined by $W =\{\{x,y\}\}$ is sufficient.
However, if $\nu$ is not minimal,  it may be that 
 $\nu={x^1, y^1, w^1_1,\dots w^1_n}$ which gives us an exponential number of possibilities for $\nu'$ while only one of them is desirable.
Intuitively, if $\nu$ is not minimal and there is some set $S$ that yields a witness for both $\nu$ and some $\nu_1\smallerNEQ\nu$,
then the set $S$ is more likely to be found when $\nu_1$ is inspected.
\end{example}

Based on this last observation, the model $\nu$ is required to be minimal.
To make the difference between $\nu$ and $\nu'$ small, and therefore make this set $S$ small, the solution $\nu'$ is required to be a maximal model. 

To obtain a minimal, respectively maximal, model from a SAT solver is done by specifying the {\em phase}---the value that the solver prefers when making decisions when traversing the search space.
Namely, preferring $0$ yields a minimal model while preferring $1$ yields a maximal model~\cite{GiunchigliaMaratea06,DiRosaEtAl10}.

\section {Computing \negationFreeAll} \label{sec:ffn-all}
\begin{algorithm}[t]
\IncMargin{1em} %
\DontPrintSemicolon%
\SetKwData{outcome}{outc}%
\SetKwData{timeout}{timeout}%
\SetKwData{initialTimeout}{initial-timeout}%
\SetKwData{success}{success}%
\SetKwData{false}{false}\SetKwData{true}{true}
\SetKwFunction{dispensable}{Free-For-Negation}%
\SetKwInOut{Input}{input}\SetKwInOut{Output}{output}
\Input{CNF formula $\phi $ and a set of variables $V$}
\Output{subset of $V$ that are free for negation}
\BlankLine

$F\gets\emptyset$\;
$X\gets V$\;
$\timeout \gets \initialTimeout $\;
\While {$X\neq \emptyset$} {
  $G\gets\emptyset $\;
  \ForEach { $x$ in $X$ } {
  $(\success,\outcome)\gets \dispensable (\phi,x,\timeout)$  \;
  \If {$\success =\true$} {
     $G\gets G\union \{x\} $\;
     \If {$\outcome =\true$} {
         $F =F\union \{x\} $\;
         $\phi =\phi\land\lnot x$
    } 
  }}
  $X\gets X\smallsetminus G $\;
$\timeout\gets k\times\timeout $\;
\Return $F$
}
\caption {Computing the set of  variables that are free for negation}\label {algo_dispensables}
\end{algorithm}

To calculate the set of  variables that are free for negation, we invoke the algorithm described in the previous section for each variable.
This procedure is optimized by conjoining the negations of the variables that have already been shown to be free for negation, which is justified by the following proposition.
\begin {proposition}\label {proposition:adding_dispensable}
Let $\phi $ and $\psi $ be formulas such that $\phi\mmodels \psi$. The formula~$\phi\land\psi$ has the same set of minimal models as $\phi$.
In particular, if $\phi\mmodels\lnot x$ then
$(\phi\land\lnot x)\mmodels\lnot y$ iff
$\phi\mmodels\lnot y$.
\end {proposition}

The motivation for conjoining negations of  variables free for
negation is to give more information to subsequent inferences.
The effectiveness of this technique, however, depends on the ordering of the variables.
Hence, the approach we use is to set timeouts for testing a single variable and if a test  times out,
the variable is tested again but with information gained from the other tests.

%
%

\autoref {algo_dispensables} summarizes these ideas in pseudocode.
The algorithm described in the previous section is represented by the function {\sf Free-For-Negation}, which returns a pair of values.
The first value in the pair indicates whether the algorithm terminated before the given timeout or not.
The second value of the pair indicates whether the given variable is free for negation or not.
The timeout is gradually multiplied by some constant coefficient $k$.
In the actual implementation there is a maximum timeout for which the algorithm stops and returns an approximation of the set of variables free for negation.

\section {Evaluation} \label{sec:res}
\autoref{algo_dispensables} was implemented in Java using SAT4j as the underlying SAT solver
while availing of its incremental interface~\cite {sat4j}. 
%
The implementation was  evaluated on a benchmark of $260$~tests\footnote {Available at \url{%
http://logos.ucd.ie/confs/jelia10/jelia10-bench.tgz}}. 
A majority of these are valid software  configurations (motivated by~\cite{JanotaEtAl10}).
A few tests are from the {SAT~'09} competition---relatively easy instances were chosen as the computed problem is significantly harder than satisfiability. 
The results appear in \autoref{tbl:eval}.  
An instance is considered solved if the answer is given in       
less than $30\,\rm s$. The time given in the table is the        
average for the solved instances. 

\begin{table}[t]\centering
\caption{Experimental evaluation}\label{tbl:eval}
\vspace {-.05cm} 
\begin{tabular}{|c|c|c|c|c|c|}
\hline
 & 
 & \multicolumn{2}{|c|}{\autoref{algo_dispensables}}
 & \multicolumn{2}{|c|}{circ2dlp+gnt} \\

 & tests
 & solved & time$[s]$
 & solved & time$[s]$ \\\hline\hline

e-shop
  & $174$
  & $174$ & $2.1$
  & $95$ & $2.4$ \\\hline

BerkeleyDB
  & $30$
  & $30$ & $0.9$
  & $30$ & $<0.1$ \\\hline

model-transf
  & $41$
  & $41$ & $1.1$
  & $35$ & $2.8$ \\\hline

SAT2009
  & $15$
  & $3$ & $7.6$
  & $2$ & $2.5$ \\\hline
\end{tabular}
\end{table}

The alternative we tried was based on the tool \texttt{circ2dlp}~\cite {OikarinenJanhunen05}, which transforms circumscription into a disjunctive logic
program, and \texttt{gnt}~\cite {DBLP:journals/tocl/JanhunenNSSY06}, which lists all models of that
program. From the list of models it is easy and fast to
construct the set of variables that are free for negation.
We also tried using a QBF solver along with~\eqref{eq:circ},
but that implementation solved \emph{none} of the $260$~tests.
\vspace {-.25cm} 
%
%
%
%


%% file: intro.tex

\section {Introduction}

Closed world reasoning (CWR) and circumscription (CIRC) are
well-known nonmonotonic reasoning techniques, that find a wide range
of practical applications. Part of the interest in these techniques is
that they bring us closer to how humans
reason~\cite{McCarthy80,Minker82,McCarthy86}.
While these techniques have been studied in the context of both
first-order and propositional logic, this paper addresses the
propositional case.
Research directions that have characterized the study of nonmonotonic
reasoning techniques include expressiveness, computational complexity,
applications and algorithms.
The different CWR rules proposed in the late 70s and 80s illustrate the
evolution in terms of expressive power in first-order and
propositional logics.
The computational complexity of propositional CWR rules was studied in
the early 90s~\cite{cadolilenzerini-AAAI90,eitergottlob-TCS93} and
showed that, with few exceptions, the complexity of CWR deduction
problems are in the second level of the polynomial hierarchy, being 
$\Pi^\text{P}_2$-complete~\cite{eitergottlob-TCS93}.
Nonmonotonic reasoning finds a wide range of applications in Artificial
Intelligence~(AI), but also in description logics~\cite{EiterEtAl-AIJ08}
and in interactive configuration~\cite{JanotaEtAl10}, among many others.
Finally, different algorithms have been proposed over the years,
examples of which include minimal model
resolution~\cite{Przymusinski-AIJ89}, tableau
calculus~\cite{Niemela-ECAI96}, Quantified Boolean Formula (QBF)
solvers~\cite{EglyEtAl00} and Disjunctive Logic Programming
(DLP)~\cite{Lifschitz96,JanhunenOikarinen04,OikarinenJanhunen05}.

The main contribution of this paper is to propose a new algorithm for
solving the deduction problem for the propositional version of some
CWR rules and for propositional circumscription.
The new algorithm is based on iterative calls to a SAT solver, and is
motivated by the practical success of modern SAT solvers.
However, given the complexity class of entailment for CWR rules, a SAT
solver can  be expected to be called an exponential number of 
times in the worst case, or be required to process an exponentially
large input.
To cope with this issue, we utilize a technique inspired in
counterexample guided abstraction refinement (CEGAR), widely used in
model checking~\cite{ClarkeEtAl-CAV00}. One of the key ideas of the
new algorithm is that we try to prove a stronger formula, which is
weakened if it turns out to be too strong. Based on this idea we
develop an algorithm that decides entailment in circumscription.
Further, we refine the algorithm to compute the closure of a formula defined by one of the variants of CWR, namely GCWA.
%
As a result, the main contributions of the paper can be summarized as
follows: 
(i) A novel algorithm for propositional circumscription that does not
require an enumeration of all minimal models or prime implicates; (ii)
Specialization of this algorithm to compute variables that are 0 in
all minimal models; and (iii) Computing the closure of GCWA.

The paper is organized as follows. The next section introduces the
notation and concepts used in the remainder of the
paper. Section~\ref{sec:probs} introduces the computational problems
addressed in the paper. The new algorithms are described in
Sections~\ref{sec:emin},~\ref{sec:ffn}, and~\ref{sec:ffn-all}. The new
algorithms are compared to alternative algorithms in
Section~\ref{sec:res}. The paper concludes with directions for future
research work.


%% file: prelim.tex

\section{Preliminaries} \label{sec:prelim}
%

%
All variables are propositional, and represented by a finite set
$V$. 
A {\em Conjunctive Normal Form} (CNF) formula $\phi$ is a conjunction of
{\em clauses}, which are disjunctions of {\em literals}, 
which are possibly negated variables.
A formula~$\phi$ can also be viewed as a set of sets of
literals. 
The two representations are used interchangeably in this
paper.
A clause is called {\em positive}, if it contains only positive literals.
Arbitrary Boolean formulas will also be considered, for which the
standard definitions apply.
A \emph{variable assignment} $\nu$ is a total function from $V$ to
$\{0, 1\}$.
In the text, a variable assignment is represented as $\{x_1^{v_1}, \dots, x_n^{v_n}\}$ 
where $V=\{x_1,\dots,x_n\}$ and $v_i\in \{0, 1\}$, $i\in 1..n$.
For a variable assignment $\nu$ and a formula $\phi$ we write
$\nu\models\phi$ to denote that $\nu$ satisfies $\phi$. In this case,
$\nu$ is called a {\em model} of $\phi$.
We write $\phi\models\psi$ if the models of $\phi$ are also models of
$\psi$.
Given a set of variables $S\subseteq V$ and $v\in\{0, 1\}$, the expression $\phi[S\mapsto
v]$ denotes the formula $\phi$ with all variables in $S$ replaced with
$v$.

\subsection{Minimal Models}

Minimal models are widely used in nonmonotonic reasoning and AI in
general.
To introduce minimal models, we consider the bitwise ordering on
variable assignments. For variable assignments $\nu$ and $\mu$ we
write $\nu \smallerLEQ \mu$ and say that  $\nu$ is {\em smaller} than
$\mu$ iff  $(\forall x\in V)(\nu(x)\leq\mu(x))$. We write
$\nu\smallerNEQ\mu$ and say that $\nu$ is {\em strictly smaller} than
$\mu$ iff $\nu\smallerLEQ\mu$ and  $\nu\neq\mu$.
A model $\nu$ of $\phi$ is a {\em minimal model } iff there is no model of $\phi$ strictly smaller than $\nu$.
%
Finally, we write $\phi\mmodels\psi$ if $\psi$ holds in all minimal
models of $\phi$.

\begin {proposition}\label {proposition:smaller}
The models of formula~$\phi$ that are strictly smaller than some variable
assignment~$\nu$ are the models of the formula
\begin {equation}\label{equation:smaller}
\phi
\land\;{%
\bigwedge%
\hspace{-.1em}\raisebox{-.5em}{$\scriptstyle\nu(x)=0$}
\;\;{\lnot x}
}\;
\land 
   {\bigvee%
     \hspace{-.45em}\raisebox{-.5em}{$\scriptstyle\nu(x)=1$}
       \;\;{\lnot x}
   } 	
\end {equation}
\end {proposition}

\subsection{Closed World Reasoning}

The intuition behind closed world assumption~(CWA) reasoning is that
facts are not considered to be true unless they were specifically stated.
This is motivated by the type of reasoning humans use on an everyday basis.
For instance, if Alice asks  Bob to buy eggs, Bob will clearly buy eggs. 
However, he  will not buy bread even though Alice has not specified that the bread should not be bought.
Traditional mathematical logic behaves differently in this respect: the fact  \textit {buy-eggs} 
trivially entails \textit {buy-eggs} but does not entail the fact $\lnot\textit{buy-bread}$.

This intuition has been realized by several different formalisms. Here
we present only a small portion of these formalisms and the interested
reader is referred to appropriate publications for further
reference~\cite{cadolilenzerini-AAAI90,eitergottlob-TCS93,DixFurbachNiemela2001}.

The standard formulation of CWA rules partitions set $V$ into three
sets: $P$, $Q$ and $Z$, where $P$ denotes the variables to be
minimized, $Z$ are the variables that can change when minimizing the
variables in $P$, and $Q$ represents all other (fixed) variables. For
any set $R$, $R^{+}$ and $R^{-}$ denote, respectively, the sets of
positive and negative literals from variables in $R$.
Following~\cite{cadolilenzerini-AAAI90,eitergottlob-TCS93}, 
a closure operation is defined for CWR rules as follows:
 
\begin {definition}
Let $\phi$ be a propositional formula, $\langle P; Q; Z\rangle$ a
partition of $V$, and $\alpha$ a CWR-rule. Then, the {\em closure} of
$\phi $ with respect to $\alpha $ is defined by,
\begin {equation}
\alpha(\phi; P; Q; Z) =\phi\cup\comprehension{\lnot K}{K
  \text { is free for negation in } \phi {\ w.r.t.\ } \alpha}
\end {equation}
\end {definition}
Each CWR rule considers a different set of formulas that are free for
negation. For each CWR rule below, a formula $K$ is free for negation
if and only if the corresponding condition holds:
\vspace {.1cm}

\cwrule{GCWA}{Generalized CWA~\cite {Minker82}}{$K$ is a positive
  literal and for every positive clause $B$ such that $\phi\nvDash B$
  it holds that $\phi\nvDash B\lor K$.}

\cwrule{EGCWA}{Extended GCWA~\cite{YahyaHenschen85-jar}}{$K$ is a
  conjunction of positive literals and for every positive clause $B$
  such that $\phi\nvDash B$ it holds that $\phi\nvDash B\lor K$.}

\cwrule{ECWA}{Extended CWA~\cite{YahyaHenschen85-jar}}{$K$ is an
  arbitrary formula not involving literals from $Z$, and for every positive
  clause $B$ whose literals belong to $P^{+}\cup Q^{+}\cup Q^{-}$, such
  that $\phi\nvDash B$, it holds that $\phi\nvDash B\lor K$.}

We consider only a subset of existing CWR rules. A detailed
characterization for existing CWR rules can be found
elsewhere~\cite{cadolilenzerini-AAAI90,eitergottlob-TCS93,DixFurbachNiemela2001}.
%
%

Observe that a single positive literal is free for negation in both
\GCWA and \EGCWA under the same conditions. Since a positive literal
corresponds to some variable, we extend the terminology for variables
accordingly.
\begin {definition}
A variable $x$ is {\em free for negation in} $\phi $ iff
for every positive clause $B $ such that $\phi\nvDash B $
it holds that $\phi\nvDash B\lor v$.
\end {definition}

Another concept closely related to closed world assumption is
circumscription. Originally, McCarthy defined circumscription in the
context of first order logic as a closure of the given theory that
considers only predicates with minimal extension~\cite
{McCarthy80}. In propositional logic, circumscription of a formula
yields a formula whose models are the minimal models of the original
one.
\begin {definition}
Consider the sets of variables $P$, $Q$ and $Z$ introduced above. The
circumscription of a formula  \mathematics {\phi} is defined as
follows:
\begin {equation}\label {eq:circ}
\circumscription (\phi;P;Q;Z) =\phi \land (\forall_{P',Z'})((\phi(P';Q;Z')\land (P'\implies P))\implies (P\implies P'))
\end {equation}
Where \mathematics {P', Z'} are sets of variables 
s.t.\ $X'=\comprehension{x'}{x\in X}$;
$\phi(P',Q,Z')$ is obtained from $\phi(P,Q,Z)$ by replacing the
variables in $P$ and $Z$ by the corresponding variables in $P'$ and
$Z'$; finally, \mathematics {P'\implies P} stands for 
\hbox{$\bigwedge_{x\in P} (x'\implies x)$}.
\end {definition}
%
%
%
In the remainder of the paper the sets $Z$ and $Q$ are assumed to be {\em
  empty}. The extension to the general case where these sets are not
empty is simple, and is outlined in Appendix~\ref{app:3set}.

It is well-known that for the propositional case, circumscription is
equivalent to ECWA~\cite{GelfondEtAl-AIJ89}.
Another well-known relationship is the one of both CWR rules and
circumscription to minimal
models~(e.g.~\cite{Minker82,cadolilenzerini-AAAI90,eitergottlob-TCS93}).
In particular variables free for negation take value 0 in all minimal
models. And, both $\EGCWA$ and circumscription entail the same set of
facts as the set of minimal models. These relations are captured by
the following propositions (adapted from~\cite{Minker82,cadolilenzerini-AAAI90,eitergottlob-TCS93}):

\begin {proposition}
A variable $x$ is free for negation in a formula $\phi $ iff $x$ is
assigned value 0 in all minimal models of $\phi $.
\end {proposition}

\begin {proposition}\label {proposition:circumscription_minimal}
Let $\phi $ and $\psi $ be formulas. 
It holds that
 $\EGCWA(\phi)\models\psi$ iff 
 $\phi\mmodels\psi$.
And, it holds that
 $\circumscription(\phi)\models\psi$ iff 
 $\phi\mmodels\psi$.
\end {proposition}

%


%% file: probs.tex

\section {Problems} \label{sec:probs}

The CWR rules yield the two following problems. The first problem
consists of computing the closure of the theory, as defined by the CWR
rule. The second problem is that of computing whether a certain fact
is entailed by that closure. 

If the closure has been computed, standard satisfiability algorithms
can be used to solve the entailment problem. However, whereas the
closure of \GCWA increases the size of the formula by at most a linear
number of literals, the closure of both \ECWA and \EGCWA may increase
the size of the formula by an exponential number of conjuncts of
literals. The circumscription of a formula can be constructed easily
but gives rise to a QBF formula and our objective is to stay within
propositional logic with the ultimate goal of developing purely
SAT-based solutions. Hence, this paper focuses on the following
problems.

\paragraph{\entailsMinimal}\mbox{}\\
\textbf {instance:} formulas $\phi $ and $\psi $\\
\textbf {question:} Does the formula $\psi $ hold in all minimal models of $\phi$?

\paragraph{\negationFree}\mbox{}\\
\textbf {instance:} formula~$\phi $ and variable~$x\in V$\\
\textbf {question:} Does $x$ take value 0 in all minimal models of $\psi$?

\paragraph{\negationFreeAll}\mbox{}\\
\textbf {instance:} formula $\phi $ and  a variable $v\in V$\\
\textbf {question:} What is the set of variables with value 0 in all minimal models of $\phi$?
\vspace{.25cm}

Note that solving \entailsMinimal enables answering whether a fact
is entailed by \ECWA or by circumscription due to \autoref
{proposition:circumscription_minimal}.  Clearly, the problem
\negationFree is a special case of \entailsMinimal with $\psi$ set to~$\lnot x$. Solving  \negationFreeAll gives us the closure
of \GCWA. 

Interestingly, in terms of complexity, the problem \negationFree  is
not easier than the problem \entailsMinimal. Both \entailsMinimal and
\negationFree are
$\Pi^\mathrm{P}_2$-complete~\cite[Lemma~3.1]{eitergottlob-TCS93}.


%% file: example1.tex
 \begin{tikzpicture}[xscale=1.6,yscale=.8]
 \node at (0, 0.3) {000};   
 \node(n100) at ( -1, 1 ) {100};   
 \node(n010) at ( 0, 1 ) { 010};   
 \node(n001) at ( 1, 1 ) { 001};

 \node(n110) at ( -1, 2 ) {110};   
 \node(n101) at ( 0, 2 ) { 101};   
 \node(n011) at ( 1, 2 ) { 011};

\draw (n110)-- node [right] {$\scriptstyle\{y\}$} (n100);
\draw (n101)-- node [below] {$\scriptstyle\{x\}$} (n001);
\draw (n011)-- node [right] {$\scriptstyle\{y\}$} (n001);
 \node at (0, 2.7 ) {111 };

\draw[dashed] (-1.25, 0)--(1.25, 0)--(1.25, 2.25)--(-1.25, 2.25)-- cycle ;  
\node [right]  at (1.25, 0) {$\scriptstyle\phi$};   

\draw (-1.3, 1.7)--(1.3, 1.7)--(1.3, 2.3)--(-1.3, 2.3)--cycle;
\node [above]  at (1.1, 2.3) {$\scriptstyle\phi\land\lnot\psi$};   
\end{tikzpicture}

%% file: conclusion.tex

\section {Summary and Future Work}
\vspace {-.2cm} 

 This paper proposes an algorithm for deduction under the set of minimal models of a propositional formula.
This algorithm enables us to reason under the propositional versions of close world assumption or circumscription. 
The algorithm hinges on an application of a SAT solver but more importantly on counterexample guided abstraction refinement (CEGAR).
While CEGAR  has been amply used in software verification~\cite {ClarkeEtAl-CAV00,FlanaganQadeer02-POPL},
we are not aware of its application in nonmonotonic reasoning.

The deduction problem under the set of minimal models can be formulated as  QBF~\cite{EglyEtAl00} or as a DLP~\cite {Lifschitz96,JanhunenOikarinen04}.
The experimental results suggest that current QBF solvers are not practical for this problem.
The comparison to  the DLP-based solution indicates that our dedicated algorithm
enables solving more instances. Nevertheless, the DLP-based solution was faster for some instances.


The promising experimental results indicate that the ideas behind the presented algorithms have potential for further work.
The evaluation was performed for the computation of variables free for negation defining the closure of a theory in GCWA, 
hence, further evaluations should be performed on other types of problems in this domain.
On a more general scale, it is well known that minimal models can be seen as optima with respect to the pertaining ordering~\cite {CastellEtAl96,DiRosaEtAl10}.
This opens possibilities to investigate generalizations of the presented algorithms for different orderings than the one used for minimal models.
Last but not least, the comparison with the DLP-based solution indicates that it would be beneficial to investigate approaches tackling the problem with hybrid techniques.


%% file: proofs_mj.tex
\section {Proofs}
\paragraph {\bf\autoref {proposition:smaller}.}
The models of formula~$\phi$ that are strictly smaller than some variable
assignment~$\nu$ are the models of the formula
\begin {equation*}
\phi\land{\bigwedge_{\nu(x)=0}\lnot x}\;\land {\bigvee_{\nu(x)=1}\lnot x} \text { \eqref{equation:smaller}}	
\end {equation*}
\begin {proof}
Let $\nu' $ be a model of \eqref {equation:smaller}. 
The assignment $\nu'$ is a model of $\phi$ because \eqref {equation:smaller} is stronger than $\phi$.
The model $\nu'$ is smaller than $\nu$ because whenever $\nu(x)=0$ holds, $\nu'(x)=0$ holds as well due to the condition $\bigwedge_{\nu(x)=0}\lnot x$.
The model $\nu'$ is strictly smaller than $\nu$ because there must be at least one variable $x$ for which 
$\nu (x) =1$ and
$\nu' (x) = 0 $
due to the condition $\bigvee_{\nu(x)=1}\lnot x $.
\qed
\end {proof}

\paragraph {\bf\autoref {proposition:main}.}
A formula $\psi$ holds in all minimal models of the formula $\phi$ iff
 any model $\nu$ of $\phi$ where $\lnot\psi$ holds is not a minimal model of $\phi$.
\[\left [\phi\mmodels\psi\right] 
 \Leftrightarrow 
\left[ 
(\forall \nu) \left((
\nu\models\phi\land\lnot\psi)
\Rightarrow (\exists \nu')(\nu' \smallerNEQ \nu \land \nu'\models \phi) \right)
\right]
\]
\begin {proof}
In classical logic, for any assignment $\nu$ either 
$\nu\models\psi$ or
$\nu\models\lnot\psi$ but not both.
Let $\phi\mmodels\psi$ and let $\nu$ be a model $\phi$ such that $\nu\models\phi\land\lnot\psi$. Then $\nu$ must not be minimal because
$\nu\models\lnot\psi$ and therefore $\nu$ would be a minimal model of $\phi$ not satisfying $\psi$.

If any model of $\phi$ that satisfies $\lnot\psi$ is not minimal,
then all the minimal models of $\phi$ must satisfy $\psi$.
\qed  
\end {proof}

\paragraph {\bf\autoref {proposition:substitution}.}
Let $\nu$ be a model of a formula $\phi$. The model $\nu$ is {\em not} a minimal model of $\phi$ 
iff
there exists a set of variables $S$ such that 
$\nu$ is a model of the formula  $\phi[S\mapsto 0]$,
and,
$\nu(x) = 1$ for some $x\in S$.
%
\begin {equation*}
(\exists \nu')(\nu' \smallerNEQ \nu \land \nu'\models \phi) 
\Leftrightarrow
(\exists S\subseteq V)\left(\nu\models \phi[S\mapsto 0]\land(\exists x\in S)(\nu(x)=1)\right)
\text { \eqref {equation:substitution}}
\end {equation*}
\begin {proof}
If $\nu$ is not a minimal model of $\phi$, then there exists some model $\nu'$ such that $\nu'\smallerNEQ \nu$.
By definition, there exists some set of variables $S$  such that  
$\nu' (x) =0$ and $\nu (x)= 1$ for  $x\in S$, and,
$\nu' (x) =\nu (x)$ for  $x\notin S$.
Then $\nu$ is a model of $\phi[S\mapsto 0]$ because $\nu'$ is a model of $\phi$ and $\nu'$ assigns $0$ to all variables in~$S$.

Let $S$ be a a set of variables such that  $\nu\models \phi[S\mapsto 0]$
and $(\exists x\in S)(\nu(x)=1)$.
Let us define the assignment $\nu'$ such that
$\nu'(x) = 0$ if $x\in S$ and
$\nu'(x) = \nu (x) $ otherwise.
Then $\nu'$ is a model of $\phi$ because $\phi [S\mapsto 0]$ corresponds to a partial evaluation of $\phi$. 
 The model $\nu'$ is smaller than $\nu$ because it differs only on the variables in $S$, where $\nu'$ is $0$.
 The model $\nu'$ is {\em strictly } smaller than $\nu$ because at least one of the variables from $S$ are assigned the value $1$ by $\nu$
due to the condition $(\exists x\in S)(\nu(x)=1)$. Hence, $\nu'\smallerNEQ \nu$ and therefore $\nu$ is not minimal.
\qed  
\end {proof}

\paragraph {\bf\autoref {proposition:dispensable_flipped}.}
Let $\nu$ be a model of a formula $\phi$ such that $\nu(x) = 1$ for a variable $x$.
If the variable $x$ is free for negation, then there exists a model $\nu'$ of $\phi$ such that
$\nu'\smallerNEQ \nu $ an $\nu'(x) = 0$.
\begin {proof}
Since  the set of considered  variables $V$ is finite, there are no infinitely decreasing chains in the ordering $\smallerNEQ$ and therefore
for the model $\nu$ there must be a model $\nu'\smallerLEQ\nu$ that is minimal. Since $x$ is free for negation, it must have the value 0 in such model $\nu'$.
\qed
\end {proof}

\paragraph {\bf\autoref {proposition:adding_dispensable}.}
Let $\phi $ and $\psi $ be formulas such that $\phi\mmodels \psi$. The formula $\phi\land\psi$ has the same set of minimal models as the formula $\phi$.
In particular, if $\phi\mmodels\lnot x$ then
$(\phi\land\lnot x)\mmodels\lnot y$ iff
$\phi\mmodels\lnot y$.
\begin {proof}
Since the formula $\psi$ holds in all minimal models of $\phi$, all the minimal models of $\phi $ are models of $\phi\land\psi$.
Since models of the formula $\phi\land\psi$ form a subset of the models of the formula $\psi$, the minimal models of $\phi$ are also minimal in $\phi\land\psi$.
To show that any minimal model of $\phi\land\psi$ is also a minimal model of $\phi$, consider for contradiction that there is an assignment $\nu$ such that
$\nu$ is a minimal model of $\phi\land\psi$ but is not a minimal model of $\phi$. 
Since $\nu$ is not minimal in $\phi$  and there are no  infinitely decreasing chains in $\smallerNEQ$, 
 there must be a minimal model $\nu'$ of $\phi$ such that $\nu'\smallerNEQ \nu$.
Since $\nu$ is minimal in $\phi\land\psi$, $\nu'$ is not a model of $\phi\land\psi$ but that is a contradiction because all minimal models of $\phi$ are  also models of~$\psi$.
\qed
\end {proof}

%% file: 3_sets.tex

\section {Deciding Entailment in Full ECWA} \label{app:3set}
The article  presents an algorithm that enables us to decide whether a formula holds in all minimal models of another formula.
This  enables us to decide entailment for ECWA  and  circumscription with $Q=Z=\emptyset$ (see \autoref {sec:prelim}).
To provide a semantic characterization supporting arbitrary $Q$ and $Z$,  the concept of minimality of models is extended.

\begin{definition}
Let $P$, $Q$, and $Z$ be a partitioning of the variables $V$.
For  variable assignments $\nu$ and  $\mu$, we write 
$\nu\smallerLEQ_{(P, Z)}\mu$ if
$\nu (x)=\mu (x)$ for all $x\in Q$, and,
$\nu (x)\leq\mu (x)$ for all $x\in P$.
We write $\nu\smallerNEQ_{(P, Z)}\mu$ if
$\nu\smallerLEQ_{(P, Z)}\mu$ and not
$\mu\smallerLEQ_{(P, Z)}\nu$. 

We write $\phi\models_{(P, Z)}\psi$ iff $\psi $ holds in all models that are minimal with respect to the ordering $\smallerNEQ_{(P, Z)}$.
\end {definition}

 The entailment  $\models_{(P, Z)}$  corresponds to  deduction from the closure defined  by ECWA  and analogously for circumscription~\cite{cadolilenzerini-AAAI90}.  So  we focus on deciding \hbox {$\phi\models_{(P, Z)}\psi$}.
Observe that $\phi\models_{(P, Z)}\psi$
coincides with $\phi\mmodels\psi$ when $Q=Z=\emptyset$.
In terms of computational complexity, deciding $\phi\models_{(P, Z)}\psi$
is not more difficult than deciding $\phi\mmodels\psi$
since both problems are $\Pi^\text{P}_2$-complete~\cite{eitergottlob-TCS93}.

We show that \autoref {algo_refinement}
can be easily modified to decide  $\phi\models_{(P, Z)}\psi$.
The structure of the algorithm remains the same, hence here we focus on the form of the abstraction and how it is refined.
Recall that the abstraction 
captures the statement that any model of $\phi$ that violates $\psi$
is not a minimal model. In particular, a smaller model can be found (see \eqref {equation:abstraction}).
The following formula replicates the same idea for the minimality defined by
$\smallerNEQ_{(P, Z)}$.

\begin {equation}\label {equation:abstraction_extended}
\begin {array} {l}
(\phi\land\lnot\psi)\Rightarrow
\bigvee_{(S,Z_0,Z_1)\in W} \left (\phi[S\mapsto 0,Z_0\mapsto 0,Z_1\mapsto 1]\land  \bigvee_{x\in S}x\right),\\
\text {\quad where } W\subseteq\comprehension {(S,Z_0,Z_1)} {S\subseteq P,Z_0\subseteq Z,Z_1\subseteq Z}
\end {array}
\end {equation}

In this case, the abstraction is defined by a set of triples
for each of the triples determines which variables  are flipped to 0
and which are flipped to 1.
Since variables from $P$ can only be flipped to 0 and variables from $Q$ cannot be flipped at all,
the right-hand side of the abstraction is indeed permitting only models smaller in the sense of 
$\smallerNEQ_{(P, Z)}$.

When the algorithm tries to refine the abstraction, it needs to find a model $\nu'\smallerNEQ_{(P,Z)} \nu$, where $\nu$ is a model of the negation of the abstraction. We observe that $\nu' $ must be a model of the following formula.
\begin {equation}
\phi\land
\bigwedge_{\nu (x) = 0\land x\in P}\lnot x\land
\bigvee_{\nu (x) = 1\land x\in P}\lnot x
\land
\bigwedge_{\nu (x) = 0\land x\in Q}\lnot x\land
\bigwedge_{\nu (x) = 1\land x\in Q}x
\end {equation}

The abstraction is refined by adding a triple into the set $W$. The triple is defined by the following elements.
\begin{equation}
\begin {array} {lcl}
S & = &\comprehension {x} {x\in P\land\nu (x) = 1\land\nu' (x) = 0}\\
Z_0 & = &\comprehension {x} {x\in Z\land\nu (x) = 1\land\nu' (x) = 0}\\
Z_1 & = &\comprehension {x} {x\in Z\land\nu (x) = 0\land\nu' (x) = 1}
\end {array}
\end {equation}